\documentclass[preprint,11pt,authoryear]{elsarticle}

\usepackage{amssymb}
\usepackage{amsthm}

\usepackage{amsmath}

\usepackage{thmtools}
\theoremstyle{definition}
\declaretheorem[numbered=no]{guidelines}

\usepackage{subcaption}

\usepackage{caption}
\usepackage[graphicx]{realboxes}
\usepackage{varwidth}

\usepackage{booktabs, rotating}

\usepackage[colorlinks]{hyperref}

\usepackage{makecell}

\usepackage{lscape}

\newcommand{\nhperit}{\ensuremath{n_1}} % number to delete 
\newcommand{\routepernh}{\ensuremath{n_2}}

\newcommand{\nhood}{\ensuremath{\eta}}

\newcommand{\feassol}{\ensuremath{\mathcal{S}}} %Set of feasible solutions
\newcommand{\sol}{\ensuremath{S}} % a solution
\newcommand{\decvarx}{\ensuremath{x}} %decision variable decide if route taken
\newcommand{\decvary}{\ensuremath{y}} %decision variable decide if route taken

\newcommand{\MLlns}{LENS}
\newcommand{\MLpotential}{\textsc{PredictPotential}}

\newcommand{\samplex}{\textbf{x}} %symbol for a list of features
\newcommand{\sampley}{y} %symbol for a list of features

\usepackage{algorithm}
\usepackage[noend]{algpseudocode}

\let\OrigComment\Comment
\renewcommand{\Comment}[1]{\OrigComment{\small#1}}
\journal{--}

\begin{document}

\begin{frontmatter}

\title{Learning-Enhanced Neighborhood Selection for the Vehicle Routing Problem with Time Windows}

\author[cwi,illc,ds]{Willem Feijen\corref{cor}}
\ead{w.feijen@cwi.nl}
\cortext[cor]{Corresponding Author}
\author[cwi,illc]{Guido Sch\"afer}
\ead{g.schaefer@cwi.nl}
\author[ds]{Koen Dekker}
\ead{koen.dekker@3ds.com}
\author[ds]{Seppo Pieterse}

\affiliation[cwi]{organization={Networks and Optimization Group, Centrum Wiskunde \& Informatica (CWI)}, %Department and Organization
            addressline={Science Park 123}, 
            city={Amsterdam},
            postcode={1098 XG}, 
            %state={},
            country={The Netherlands}}
\affiliation[illc]{organization={Institute for Logic, Language and Computation (ILLC), University of Amsterdam}, %Department and Organization
            addressline={Science Park 107}, 
            city={Amsterdam},
            postcode={1098 XG}, 
            %state={},
            country={The Netherlands}}
\affiliation[ds]{organization={DELMIA R\&D, Dassault Systèmes B.V.}, %Department and Organization
            addressline={Utopialaan 25}, 
            city={'s-Hertogenbosch},
            postcode={5232 CD}, 
            %state={},
            country={The Netherlands}}       
\begin{abstract}
%% Text of abstract
Large Neighborhood Search (LNS) is a universal approach that is broadly applicable and has proven to be highly efficient in practice for solving optimization problems.
We propose to integrate machine learning (ML) into LNS to assist in deciding which parts of the solution should be destroyed and repaired in each iteration of LNS. 
We refer to our new approach as \emph{Learning-Enhanced Neighborhood Selection} (\emph{\MLlns} for short). 
Our approach is universally applicable, i.e., it can be applied to any LNS algorithm to amplify the workings of the destroy algorithm.
In this paper, we demonstrate the potential of \MLlns\ on the fundamental Vehicle Routing Problem with Time Windows (VRPTW). 
We implemented an LNS algorithm for VRPTW and collected data on generated novel training instances derived from well-known, extensively utilized benchmark datasets. 
We trained our LENS approach with this data and compared the experimental results of our approach with two benchmark algorithms: a random neighborhood selection method to show that LENS learns to make informed choices and an oracle neighborhood selection method to demonstrate the potential of our LENS approach.
With LENS, we obtain results that significantly improve the quality of the solutions.
\end{abstract}

% \begin{highlights}
% \item Present Learning Enhanced Neighborhood Selection (\MLlns) for Large Neighborhood Search
% \item \MLlns\ can be applied to any LNS algorithm to amplify workings of the destroy algorithm
% \item We demonstrate the potential of \MLlns\ on the Vehicle Routing Problem with Time Windows
% \item With LENS, we obtain results that significantly improve the quality of the solutions
% \end{highlights}


\end{frontmatter}

\section{Introduction}

\paragraph{Motivation and Background}

Efficiency in the planning of large logistics companies is of major importance, both for reducing the environmental footprint and for reducing costs. To create efficient schedules, planners at those companies need to be able to obtain high-quality solutions quickly. In practice, such solutions are often computed by using an \emph{iterative approach} that proceeds along the following lines: (1) create an initial feasible solution, (2) iteratively improve upon the current solution until a certain termination criterion is met (e.g., maximum number of iterations is reached, sufficient solution quality is achieved).
\emph{Large Neighborhood Search (LNS)} is one such approach that is broadly applicable and has proven to be highly efficient in practice (see, e.g., the survey by \citet{WINDRASMARA2022105903}).
Each iteration of LNS consists of two steps: the \emph{destroy method} and the \emph{repair method}.
For the repair method, often a general-purpose solver like a mixed integer programming (MIP) solver or a constraint programming solver can be used \citep{pisinger2010large}. If available, one can also take advantage of heuristics that are known to be well-performing and embed them as a repair method \citep{WINDRASMARA2022105903}.
However, for the destroy method the situation is different: often this asks for the development of specialized algorithms tailored towards the specific scheduling problem, which requires expert knowledge and needs time to be developed and implemented.

As also mentioned by \citet{Ropke2006Adaptive}, LNS works particularly well if the considered problem can be easily partitioned into a number of subproblems where some constraints must be satisfied, covered by a master problem to control how the subproblems are combined. Many routing and scheduling problems conform to this structure and are therefore oftentimes successfully solved with LNS in practice. 
A recent survey paper by \citet{WINDRASMARA2022105903} on a specific type of LNS, namely \emph{Adaptive Large Neighborhood Search (ALNS)}, categorized 252 scientific publications. Most of these articles are on routing and scheduling, but there are also applications in, e.g., manufacturing and agriculture. Below, we highlight a few specific applications of LNS to show its broad applicability.

\citet{rastani2021large} use LNS to solve the \emph{Electric Vehicle Routing Problem with Time Windows}, where a level of complexity is added to standard \emph{Vehicle Routing Problem with Time Windows (VRPTW)} since vehicle batteries need to be recharged during the day. Factors like temperature and speed are taken into account to calculate a battery's range, and in this paper specifically, the carried load of a vehicle is taken into account. \citet{chen2021adaptive} solve another variant of VRPTW, named \emph{Vehicle Routing Problem with Time Windows and Delivery Robots}. In this variant, vehicles can carry multiple robots which can take over part of the deliveries. An LNS algorithm is used to solve this problem. In a study by \citet{kuhn2021integrated}, the Vehicle Routing Problem is intertwined with the order batching problem, where multiple customers are grouped together. They solve it using \emph{General Adaptive Large Neighborhood Search}, a newly introduced Adaptive LNS method inspired by \emph{General Variable Neighborhood Search} (see \citep{hansen2019variable}).

\paragraph{Our Approach: Leaning-Enhanced Neighborhood Selection (\MLlns)}

\medskip\noindent
In this paper, we propose to integrate machine learning (ML) into LNS to assist in deciding which parts of the schedule should be destroyed and repaired in each iteration. 
In particular, we investigate how to exploit machine learning techniques to amplify the workings of any possible destroy algorithm. 
Conceptually, our new approach can be applied to any LNS that makes some random choices (explained in more detail below). We refer to our new approach as \emph{Learning-Enhanced Neighborhood Selection} (\emph{\MLlns} for short).
%We applied our ideas to the Vehicle Routing Problem with Time Windows. 

This research was inspired by experimental findings that we obtained for a real-world application solving large-scale routing problems on a daily basis. When using our \MLlns\ approach to guide the destroy method of a (highly sophisticated) LNS for this application we observed that this leads to a significant speed-up of the optimization process. 

The \MLlns\ approach described in this paper is based on similar ideas. However, our approach is implemented for and tested on publicly available and well-studied benchmark instances of the \emph{Vehicle Routing Problem with Time Windows} (see also \cite{accorsi2022Guidelines}).
The algorithms and data sets that were used for the experiments reported in this paper are publicly available from the following repository:

\medskip
\centerline{\url{https://github.com/w-feijen/ML4LNS}}

\bigskip
We remark that the improvements of our \MLlns\ approach that we observed for the real-world application are (significantly) better than the ones we can report for VRPTW here.\footnote{Unfortunately, due to a non-disclosure agreement, we are unable to share our experimental findings for the real-world application.} 
We comment on this in more detail in the conclusions. 

\newpage
\paragraph{Our Contributions}
The main contributions presented in this paper are as follows:

\begin{enumerate}
    \item %Idea of using ML in LNS. 
    We introduce a general \emph{Learning-Enhanced Neighborhood Selection approach, referred to as \emph{\MLlns} for short}, that integrates ML in the destroy method of an LNS algorithm. This approach is based on supervised learning and can be used as an enhancement of the workings of any destroy method 
    using some form of randomization (as explained below).
    
    \item We create an LNS algorithm for the VRPTW. The algorithm consists of (1) a destroy heuristic that exploits a newly defined distance measure, based both on distance and time windows, and (2) an existing repair heuristic.
    
    \item  We apply our \MLlns\ approach to the LNS algorithm of VRPTW. For this purpose, we first define features that describe the potential improvement of a set of routes. We then collect data on these features and build an ML model on top of it. Our ML model can predict whether or not an improvement can be found in a given set of routes.

   \item We provide general guidelines on how to collect the right data sets if supervised learning-based ML is used in optimization algorithms. Based on our experiments, it seems crucial to perform multiple iterations of data collection, where a premature ML model is used to guide subsequent data collection iterations. 
   Eventually, the final ML model is trained on all the collected data.
   By following these guidelines, we ensure that the relevant information is gathered on runs in which the optimization is guided by the ML approach.
    
   \item We generate a training set of VRPTW instances based on the \emph{R1 instances} of \citet{homberger2005two} consisting of 1000 customers. 
   Our generated training instances can be used to collect data samples for future supervised learning studies on the R1 instances.
    
\end{enumerate}

\paragraph{Related Work}
Combining machine learning with optimization is a hot topic. There is a survey about reinforcement learning in combinatorial optimization by \citet{mazyavkina2021reinforcement}, about enhancing optimization algorithms with ML for Vehicle Routing Problems by \citet{bai2023Analytics} and an overview which distinguishes three different paradigms of combinations between ML and combinatorial optimization by \citet{bengio2021machine}. The first of these paradigms is to leverage machine learning to solve combinatorial optimization problems directly from the input instance. As opposed to this first paradigm, in which the combinatorial algorithm is replaced by an ML algorithm, the second and third paradigms use ML next to combinatorial methods, either in a single place (paradigm 2), or repetitively (paradigm 3).

\citet{accorsi2022Guidelines} give very useful guidelines for testing ML approaches in Vehicle Routing Problems (VRP). We highlight some of their advice: first, they stress the importance of a clear problem description and the representativeness of the test instances. In particular, for VRPTW they advise using the instances by \citet{homberger2005two}.
Moreover, they stress including the best available algorithms to benchmark against, and they give examples of how to visualize comparisons between algorithms. 

Moreover, \citet{accorsi2022Guidelines} and also \citet{wu2019learning} state that most of the proposed approaches consider \emph{construction heuristics}, in which ML is used to construct a feasible solution. 
An example of using ML to construct a feasible solution is a study by \citet{Kool2019Attention}, in which a solution to the Travelling Salesperson Problem (TSP) is constructed iteratively.
A few others focus on using ML in \emph{improvement heuristics}, to guide the exploration of the search space and iteratively improve an existing solution. 
We focus on these improvement heuristics, which follow the third paradigm by \citet{bengio2021machine} of combining combinatorial optimization and ML and replacing some of the intermediate steps in the larger improvement heuristic framework. Some related works in which ML is used to enhance search heuristics are given by \citet{wu2019learning, hottung2020neural, li2021learning, sonnerat2021learning}. We elaborate on these examples below.

\citet{wu2019learning} use reinforcement learning in a neighborhood search to solve TSP and VRP without time windows. In particular, RL decides which pair of nodes to feed to a pairwise improvement operator. Similarly to \citet{wu2019learning}, we use ML to choose what to destroy in the solution. On the contrary, we choose a substantially larger part to destroy. \citet{hottung2020neural} augment an LNS algorithm with a neural network to solve capacitated VRP without time windows. However, the neural network model is not used in the destroy step but in the repair step. \citet{li2021learning} augment a local search algorithm for VRP without time windows with a transformer network, that identifies which set of routes needs to be optimized. A black box solver is used in the improvement step. Finally, \citet{sonnerat2021learning} use an LNS algorithm to solve Mixed Integer Programs and ML is used to decide which of the variables to destroy. 

Another line of research worth mentioning is the \emph{Adaptive Large Neighborhood Search (ALNS)}, introduced by \citet{Ropke2006Adaptive} which is an extension of LNS. In each iteration, ALNS makes a choice on which of several destroy or repair methods to use, based on weights that are adapted during the run of the algorithm. A meta-analysis on the impact of the additional adaptive layer is done by \citet{turkevs2021meta}. It shows that the adaptive layer improves the objective function value by 0.14\%. Since the adaptive layer does add extra complexity, they recommend it only in some specific situations.
\section{Preliminaries}

We introduce the following notation to describe a generic combinatorial optimization problem. An instance $I = (\feassol, c)$ of a combinatorial optimization problem is given by a set $\feassol$ of feasible solutions of $I$ and a cost function (or objective function) $c:\feassol\mapsto \mathbb{R}$ which defines a cost $c(\sol)$ for every solution $\sol \in \feassol$. 
Without loss of generality, we consider a minimization problem and the goal is to find a feasible solution $\sol^*\in \feassol$ such that $c(\sol^*) \leq c(\sol)$ for all $\sol\in \feassol$. We call $\sol^*$ an \emph{optimal solution} of $I$.

\paragraph{Large Neighborhood Search (LNS)}

Our algorithm is based on an iterative local search approach, known as \emph{Large Neighborhood Search (LNS)} (see, e.g., \citet{pisinger2010large} and \citet{shaw1998using}). Note that LNS is a universal approach that can be applied to any generic combinatorial optimization problem. We describe the approach in more detail below (see also Algorithm \ref{LNS}).

Let $I = (\feassol, c)$ be an instance of the optimization problem under consideration. 
LNS then starts with an arbitrary feasible initial solution $\sol \in \feassol$ as input.
The algorithm keeps track of the best solution $\sol^{\text{best}}$ encountered so far. 
In each iteration, a (small) part of the solution $\sol$, called \emph{neighborhood}, is selected, which is then destroyed and repaired (or rebuilt) again. 
The former and latter are done by a so-called \emph{destroy method} and \emph{repair method}, respectively.
Crucially, the repair method only rebuilds a small part of the (potentially very large) solution, and might therefore outperform an approach that re-optimizes the whole solution. The goal of alternating the destroy and repair operations is to compute a solution that improves the objective function value. 
With this aim, an \emph{accept method} is used to determine whether the improvement of the newly created solution is significant enough (e.g., in terms of a decrease in objective function value) to be used subsequently. 
Furthermore, the best-known solution $\sol^{\text{best}}$ is updated if necessary.
The algorithm continues this way until a predefined \textsc{StoppingCriterion} is met. 

As a key concept, we introduce our notion of a \emph{neighborhood}, denoted by \nhood\ in Algorithm \ref{LNS}. 
Typically, a solution \sol\ is defined by a (possibly ordered) set of solution elements. For a given solution, we define a {neighborhood} as a subset of these elements that might be selected in order to be destroyed.\footnote{We remark that our notion of a neighborhood differs slightly from the one typically used in the context of LNS, where it refers to the set of all solutions that can be obtained from a given solution by applying the destroy and repair methods.}
It remains problem-specific how these neighborhoods are defined precisely.
For example, for a routing problem like VRPTW the neigborhood could be a subset of routes or customers (see also below), and for a scheduling problem it could be a subset of machines or jobs.

\begin{algorithm}[t]
    \caption{\textsc{Large Neighborhood Search}}
    \label{LNS}
    \begin{algorithmic}[1] % The number tells where the line numbering should start
        \State \textbf{Input:} feasible solution $\sol\in \feassol$
        \State $\sol^{\text{best}}=\sol$
        \Repeat
            \State $\nhood = \textsc{SelectNeighborhood}(\sol)$ \label{alg:lns:ns}
            \State $\sol^{\text{temp}}=\textsc{Repair}(\textsc{Destroy}(\sol, \nhood))$   
            \If{\textsc{Accept}$(\sol^{\text{temp}},\sol)$}
                $\sol = \sol^{\text{temp}}$\EndIf
            \If{$c(\sol^{\text{temp}}) < c(\sol^{\text{best}})$}
                 $\sol^{\text{best}} = \sol^{\text{temp}}$
            \EndIf
        \Until{\textsc{StoppingCriterion} is met}
        \State \Return $\sol^{\text{best}}$
    \end{algorithmic}
\end{algorithm}

\bigskip
The LNS algorithm does not specify how the respective stopping criterion and the neighborhood selection, destroy, repair, and accept methods are defined, as this is problem-specific. 
The destroy method simply destroys the neighborhood that it gets as input.
In practice, many advanced and well-working procedures are used for the implementation of the repair method, e.g., integer linear programming solvers or advanced path-building algorithms.
The accept method could be a simple hill-climbing procedure, in which only improving solutions are accepted, but also simulated annealing can be used. A stopping criterion could be based on, e.g., time, number of iterations, or the value of the best-known solution. A novel neighborhood selection method is given in this paper.%
\section{Learning-Enhanced Neighborhood Selection (LENS)}\label{ss:destroy}

\citet{pisinger2010large} observe 
that a destroy method typically uses randomization to ensure that different parts of the solution (i.e., neighborhoods) are destroyed in each iteration. 
However, if neighborhoods are destroyed that were already of high quality, it might be hard to find any improvement by repairing the solution again subsequently. Instead, one would rather want to destroy neighborhoods
in which much improvement can be found. 

Based on this observation, we propose to use information from past iterations to make a decision on which neighborhoods
to destroy. 
We do this by using ML techniques to predict the improvement gained after destroying and repairing a certain neighborhood.
Ideally, the prediction enables us to identify a low-quality part of the solution such that the objective value improves significantly when this part is destroyed and repaired.
Algorithm \ref{Destroy-learn} contains the pseudocode for our proposed \emph{Learning-Enhanced Neighborhood Selection (LENS)} method, which can be used as a \textsc{SelectNeighborhood} routine in Line \ref{alg:lns:ns} of LNS (Algorithm \ref{LNS}). 

\begin{algorithm}[t]
    \caption{\textsc{Learning-Enhanced Neighborhood Select.~(\MLlns)}}
    \label{Destroy-learn}
    \begin{algorithmic}[1] % The number tells where the line numbering should start
        \State \textbf{Input:} feasible solution $\sol \in \feassol$ and integer \nhperit
        \For{$j = 1,\ldots,\nhperit$}
            \State $\nhood_j = \textsc{CreateNeighborhood}(\sol)$
            \State $p_j = \MLpotential(\sol,\nhood_j)$
        \EndFor
        \State $j^* = \arg\max_j {p_j}$
        %\State $ \sol = $ \textsc{Destroy}$(\sol,\nhood_{j^*})$
        \State \Return $\nhood_{j^*}$
    \end{algorithmic}
\end{algorithm}

\MLlns\ starts with creating \nhperit\ candidate neighborhoods to destroy. 
More specifically, the \textsc{CreateNeighborhood} method identifies a candidate part $\nhood_j$ of the solution to destroy.
After that, an ML procedure $\textsc{PredictPotential}$ predicts the potential $p_j$ (in terms of improvement) of each candidate neighborhood $\nhood_j$. 
The neighborhood with the highest potential is selected and returned.  
Subsequently, the returned neighborhood will be destroyed and repaired in the LNS algorithm.

Ideally, we would want to generate a diverse pool set of $n$ candidate neighborhoods to choose from. This will be guaranteed if the \textsc{CreateNeighborhood} method uses randomization to create a new neighborhood. Generally, our approach applies whenever there is a meaningful way to generate such a pool set of neighborhoods. 

\paragraph{Machine Learning Model}%\label{ss:mlMOdel}

We describe the ML model that we use to predict the potential in our LENS approach described above (see Algorithm \ref{Destroy-learn}). We propose to construct a supervised classification model. 
We need to train this supervised classification model, and therefore we need to create a history of previous iterations that we can learn from. 

The first step for creating this history is to define features. These features describe the neighborhood and are used by the ML model to make a prediction. Defining the right features is an application-specific problem.  %and is therefore not elaborated on here. 
We give an example of features for the application to VRPTW in Section \ref{ss:vrptw:destroy} and in \ref{appendix:1}. 
A sample in the history consists of 
(1) a set of features describing the neighborhood, and 
(2) the improvement that this neighborhood gives.
We denote a sample as follows: for the $i$'th iteration, and the $j$'th neighborhood in that iteration ($j \in \{1, \ldots, \nhperit \}$), the sample is denoted by $(\samplex_{ij}, \sampley_{ij})$.

\begin{algorithm}[t]
    \caption{\textsc{Data Collection}}
    \label{Destroy-datacollection}
    \begin{algorithmic}[1] % The number tells where the line numbering should start
        \State \textbf{Input:} feasible solution $\sol \in \feassol$ and integer \nhperit\ %and iteration index $i$
        \State $i=0$
        \Repeat
            \State $i = i +1$
            \For{$j = 1,\ldots,\nhperit$}
                \State $\nhood_j = \textsc{CreateNeighborhood}(\sol)$
                \State $\samplex_{ij} = \textsc{ComputeFeatures}(\sol, \nhood_j)$
                \State $\sol_j^{\text{\text{temp}}} = \textsc{Repair}(\textsc{Destroy}(\sol, \nhood_j))$ 
                \State $\sampley_{ij} = \max(c(\sol)- c(\sol_j^{\text{temp}}),0)$
                \State store $(\samplex_{ij}$, $\sampley_{ij})$
            \EndFor
            \If{ \textsc{DCStrategy} is random}
                \State $j^* = \textsc{RandomInteger}(1,\nhperit)$
            \Else \Comment{\textsc{DCStrategy} is ML}
                \State $j^* = \arg\max_j \MLpotential(\sol,\nhood_j)$
            \EndIf%
    %        % \State $ \sol = $ destroy$(\sol,\nhood_{i^*})$
            %\State $\sol^{\text{temp}}= \sol_{j^*}^{\text{temp}}$
            \If{\textsc{Accept}$(\sol_{j^*}^{\text{temp}},\sol)$}
                $\sol = \sol_{j^*}^{\text{temp}}$\EndIf
        \Until{\textsc{StoppingCriterion} is met}
    \end{algorithmic}
\end{algorithm}

Our algorithm \textsc{Data Collection} creates a history by collecting such samples (see Algorithm \ref{Destroy-datacollection}). 
We elaborate on Algorithm \ref{Destroy-datacollection} in more detail. 

In each iteration of the \textsc{Data Collection} algorithm, the \textsc{CreateNeighborhood} method creates $\nhperit$ neighborhoods. 
As opposed to Algorithm \ref{Destroy-learn}, the repair method is executed for \emph{all} of these neighborhoods. However, none of the updated solutions given by the repair method are directly accepted.
Instead, in the $i$'th iteration of the \textsc{Data Collection} algorithm, both the features that describe the $j$'th neighborhood and the improvement that is gained after destroying and repairing this neighborhood are stored in the sample $(\samplex_{ij}$, $\sampley_{ij})$.
After storing the $\nhperit$ samples for iteration $i$, one neighborhood is selected. 
{\sol\ is updated if }the repaired solution for this neighborhood is accepted and the {data collection} {LNS} continues with the next iteration.

Which neighborhood is selected is based on the \emph{data collection strategy} (\textsc{DCStrategy} in Algorithm \ref{Destroy-datacollection}). 
This strategy can either be random or follow predictions from a given ML model.
During the first run of data collection, we use a random strategy, meaning one of the computed neighborhoods will be chosen uniformly at random.
{For an ML model to generalize well, it is crucial to learn from data that comes from the same distribution as the data used for testing.}
%If data is collected with the random data collection strategy only, then no data is collected about the search space encountered when neighborhood selection is guided by ML, as done in LENS.
An ML model based on the data from random data collection only will probably not work well, since it guides the LNS to solutions on which no samples were collected during data collection.
Therefore, it is important to collect data on the search space encountered when neighborhood selection is guided by ML, as done in our \MLlns\ algorithm.
This is done by executing subsequent runs of data collection, this time guided with an ML strategy. 
After each run of data collection, a new ML model is trained on all previously collected data, which is then used to guide the subsequent run.
By doing so, the idea is that each new ML model is trained on data that has a better resemblance to the data that will be encountered during testing. 

To summarize, we give the following guidelines:
\begin{guidelines}\label{guideline} \mbox{}
\begin{enumerate}
\item Perform data collection with random {data collection strategy}. 
\item Based on the collected data, create an ML model, ML$k$ with $k = 1$.
\item Given ML$k$ for some $k \ge 1$,
%{Let $k$ be the index such that ML$k$ is the latest ML model. 
perform another run of data collection with ML$k$ as the new {data collection strategy}.
\item Create a new ML model ML$(k+1)$, based on collected data in all previous runs of data collection.
\item Repeat steps 3 and 4 to create new ML models ML3, ML4 and ML5.
\end{enumerate}
\end{guidelines}

The used ML models are classification models, and therefore the samples need to be labeled before they can be used for training. 
We define an improvement threshold and each data sample $(\samplex_{ij}, \sampley_{ij})$ gets label 1 if its improvement $\sampley_{ij}$ is above the threshold; otherwise, it gets label 0.

When the ML model is used in our \textsc{LENS} algorithm, neighborhoods are ranked based on the output of the ML model. Therefore we need to choose an ML model that can output probabilities, like a neural network or a random forest. 

\section{Application to VRPTW}
We applied our \MLlns\ technique to the VRPTW, which we explain in the following section.

\subsection{VRPTW Definition}
The vehicle routing problem is defined on a complete (undirected) graph $G=(\mathcal{V},E)$ on $n+2$ nodes. Without loss of generality, we identify the node set $\mathcal{V}$ with $\{0, 1, \dots, n+1\}$. There are two designated nodes, indicated by 0 and $n+1$, that are called the \emph{starting} and \emph{ending depot}, respectively. 
We assume that the starting and ending depot are the same (i.e., they are co-located), but conceptually it is more convenient to distinguish between them in the formulation (as will become clear below). 
The remaining nodes correspond to customers requiring service and we use $\mathcal{V}^{CST} = \{1, \dots, n\}$ to refer to this set (excluding the depots).
Each edge $(i,j)\in E$ represents a possible trip between node $i$ and $j$ with associated travel time $d_{ij}$. 
{Oftentimes, as is also the case for the test instances that we consider in this paper, customer locations are defined by coordinates on a grid and travel times are represented by the Euclidean distance between the respective locations.}
Each customer needs to be visited by one of the $m$ identical vehicles in the fleet. The vehicles all start and end at the starting and ending depot, respectively. 

In the \emph{Vehicle Routing Problem with Time Windows (VRPTW)}, extra capacity and time window constraints are imposed on the solution. More specifically, each customer $i$ has a non-negative demand $q_i$ which needs to be fulfilled by the vehicle, whilst the maximum capacity of each vehicle is limited to $Q$. For the time window constraints, each customer is associated with a service duration $\tau_i$ and an interval $[e_i, l_i]$, called the \emph{time window}, in which the service must start. Note that an arrival at a customer before $e_i$ is allowed, but will force the vehicle to wait until $e_i$ before the customer can be served. Moreover, the end of the time window specifies the latest starting time of the service; in particular, a service that starts before $l_i$ but ends after $l_i$ is allowed. For the depots, we define $q_0=q_{n+1}=0$ and $\tau_0=\tau_{n+1}=0$. The time window of the depots, $[e_0,l_0] = [e_{n+1}, l_{n+1}]$, defines the entire time horizon of the problem during which all customers have to be served. 

There are different objective functions studied in the context of VRPTW. Here, we consider the case in which we want to minimize the total travel distance of the vehicles over the set of all feasible solutions that satisfy all customer requests (i.e., demand and time window) by using at most $m$ vehicles of capacity $Q$.

The following is a Mixed Integer Programming (MIP) formulation of the problem (see, e.g., the formulation by \citet{munari2016generalized}).
\begin{alignat}{4}
\min \sum_{i\in \mathcal{V} } \sum_{j\in \mathcal{V} } d_{ij}\decvarx_{ij}&&& \label{mip:obj}\\
\text{s.t.} \sum_{j\in \mathcal{V}\setminus\{0\} } \decvarx_{ij} & = 1 && \forall i \in \mathcal{V}^{CST} \label{cons:departfromi}\\
\sum_{i\in \mathcal{V}\setminus\{{n+1}\}}\decvarx_{ij} & = 1 && \forall j \in \mathcal{V}^{CST} \label{cons:arriveatj}\\ 
\sum_{j\in \mathcal{V}\setminus\{0\}}\decvarx_{0j} & \leq m  \label{cons:mvehicles}\\
\decvary_j \geq \decvary_i + q_j \decvarx_{ij} - &Q (1-\decvarx_{ij}) \quad && \forall i, j  \in \mathcal{V} \label{cons:cumcapacity}\\
q_i \leq \decvary_i &\leq Q \quad && \forall i \in \mathcal{V} \label{cons:capacity}\\
t_j \geq t_i + (d_{ij}+\tau_i)\decvarx_{ij} &- (l_0-e_0) (1-\decvarx_{ij}) \quad && 
\forall i, j \in \mathcal{V},\ i \neq n+1,\ j \neq 0 \label{cons:cumtiming} \\
e_i \leq t_{i} &\leq l_i && \forall i\in \mathcal{V} \label{cons:timing}\\
\decvarx_{ij}& \in \{0,1\} && \forall i, j \in \mathcal{V} \label{cons:xinteger}\\
t_{i} & \geq 0, \quad y_i \in \mathbb{R} && \forall i\in \mathcal{V} \label{cons:treal}
\end{alignat}

\noindent 
There are three different types of decision variables in this MIP formulation: 
\begin{itemize}
    \item $\decvarx_{ij}$ indicate if customer $j$ is visited immediately after $i$
    %, for $i\in \mathcal{V}$ and $j\in \mathcal{V}$,
    \item $t_{i}$ is the service time of customer $i$
    %, for $i\in \mathcal{V}$,
    \item $\decvary_i$ is the cumulative demand on the route that visits customer $i$ up to and including this visit 
    %, for $i\in \mathcal{V}$. 
\end{itemize}
Constraints \eqref{cons:departfromi} and \eqref{cons:arriveatj} make sure that each customer is visited exactly once. Constraint \eqref{cons:mvehicles} ensures that at most $m$ vehicles leave the depot. The vehicle capacities are monitored by constraints \eqref{cons:cumcapacity} and \eqref{cons:capacity}. %\gs{Explain! Usage of $y_i$ too implicit!} 
{Constraint \eqref{cons:cumcapacity} ensures that the cumulative capacity is increased from $y_i$ to $y_i+q_j$, if customer $j$ is visited immediately after $i$ and constraint \eqref{cons:capacity} makes sure that the maximum capacity of a vehicle is not violated. }
Constraint \eqref{cons:cumtiming} and \eqref{cons:timing} deal with the time window restrictions. More specifically, constraint \eqref{cons:cumtiming} ensures that if customer $j$ is served after $i$, then the service time of $j$ is at least the service time of $i$ plus the distance from $i$ to $j$ plus the service duration at $i$. This constraint also eliminates sub-tours. Constraint \eqref{cons:timing} makes sure that the arrival is within the time window. 

Recall that we use $I = (\feassol, c)$ to denote an instance of the combinatorial optimization problem under consideration, where $\feassol$ is the set of all feasible solutions and $c:\feassol\mapsto \mathbb{R}$ defines a cost $c(\sol)$ for every solution $\sol \in \feassol$. For VRPTW, we adopt the convention that a feasible solution is represented by a set $\sol:= \{ r_1, r_2, \ldots, r_m\}$ of at most $m$ routes,  where each $r_i$ is an ordered set of customers from $\mathcal{V}^{CST}$. The cost $c(\sol)$ then simply refers to the total distance of all routes in $\sol$.

\subsection{Large Neighborhood Search for VRPTW}

In order to solve the VRPTW problem with LNS we need to define how to obtain an initial feasible solution, the \textsc{Destroy}, \textsc{Repair} and \textsc{Accept} methods, and a \textsc{StoppingCriterion}.
Below, we elaborate in more detail on our 
\textsc{Destroy} method (Section \ref{ss:vrptw:destroy}) and \textsc{Repair} method (Section \ref{ss:vrptw:repair}). 
To obtain an initial solution, we use an open-source optimization engine for vehicle routing problems called VROOM (see \citep{vroom_v1.13}). 
That is, we simply call VROOM to compute an initial solution for each of our VRPTW test instances.
For the \textsc{Accept} method, we use a simple \emph{hill-climbing procedure}, i.e., we only accept a new solution if its distance is smaller than the current one. Our \textsc{StoppingCriterion} is such that the LNS algorithm stops after a fixed number of iterations.

\subsubsection{\textsc{Destroy} Method}
\label{ss:vrptw:destroy}

To use the {\MLlns\ method}
(see Section \ref{ss:destroy}) we need to define a \textsc{CreateNeighborhood} method and define features that will be used to make an ML prediction. 

Our \textsc{CreateNeighborhood} method is given in Algorithm \ref{alg:createnhood}. First, we choose a so-called \emph{anchor route} $r_\textrm{a}$ by choosing one route uniformly at random from the set of routes $\sol$ in the current solution. Then, the neighborhood will be created around this anchor route {as follows}. Let $\sol^o = \sol \setminus \{r_a\}$ denote the set of remaining routes.
We sort the routes in $\sol^o$ based on a specific distance measure $\tilde{d}(\cdot, \cdot)$ between routes (defined below).
We relabel the routes in $\sol^o$ such that after the sorting it holds that for $r_i, r_j \in \sol^o$:
$$
i \leq j \iff \tilde{d}(r_{\textrm{a}},r_i) \leq \tilde{d}(r_{\textrm{a}},r_j).
$$
Based on this ordering, we define rank-based probabilities $\textsc{RBP}(\cdot, D)$, depending on some parameter $D$, that ensure that routes with a smaller distance to the anchor route $r_a$ have a higher probability to being added to the neighborhood. 
More specifically, $\textsc{RBP}(\sol^o, D)$ defines a probability $p_i$ for each route $r_i$ in the ordered set $\sol^o$ as follows: 

$$
p_i = \frac{\bar{p}_i}{\sum_{i: r_i \in \sol^o} \bar{p}_i}, \text{ where } \bar{p}_i = (|\sol^o|- i )^D.
$$ 

{Here $D>1$ is a parameter that controls how much the probabilities differ from each other. 
A large $D$ results in probabilities that are far apart and, consequently, the random sample drawn in Line \ref{alg:createnhood:sample} of  Algorithm \ref{alg:createnhood} will almost surely return the first \routepernh\ available routes. On the contrary, a small value for $D$ will result in probabilities that are closer together, and therefore cause more variety in the outcome of the random sample.}

We use this probability distribution $p = (p_1, \dots, p_{m-1})$ over the routes in $\sol^o$ to sample $\routepernh$ additional routes. Those \routepernh\ routes then form the neighborhood together with the anchor route.

%We define the distance between two routes as the distance between their centroids. 
\begin{algorithm}[H]
    \caption{\textsc{CreateNeighborhood}}
    \label{alg:createnhood}
    \begin{algorithmic}[1] % The number tells where the line numbering should start
        \State \textbf{Input:} set of routes $S$ and integer $n_2$
        \State Let $r_{\textrm{a}} = \textsc{RandomRoute}(\sol)$ \Comment{Choose anchor route uniformly at random}
        \State $\sol^o = \sol \setminus r_{\textrm{a}} $  \Comment{All other routes}
        \State \textsc{relabel}$(\sol^o, r_{\textrm{a}})$ 
        \Comment{Sort $\sol^o$ according to distance to $r_{\textrm{a}}$}
        %\State relabel $r_i \in \sol\setminus \{r_{a}\}$ according to $\tilde{d}(r_{a},\cdot)$ such that $i \leq j \iff \tilde{d}(r_{a},r_i) \leq \tilde{d}(r_{a},r_j) $
        \State $p = (p_1, \dots p_{m-1}) = \textsc{RBP}(\sol^o, D)$ \Comment{Compute rank-based probabilities}
        \State $\eta = r_{\textrm{a}} \cup \textsc{Sample}(\sol^o, \routepernh, p)$ \Comment{Sample \routepernh\ routes using $p$} \label{alg:createnhood:sample}
        \State \Return $\eta$
    \end{algorithmic}
\end{algorithm}

We elaborate in more detail on the definition of our distance measure $\tilde{d}(r_i, r_j)$ used above. Ideally, the distance measure is not based on customer locations only but also takes differences in time windows into account. 
Especially tight time windows impact how interchangeable the customers are between routes.

The distance from route $r_i$ to $r_j$ is defined as follows: 
$$ \tilde{d}(r_i, r_j) := \min_{u\in r_i} \{ \text{dist}(u,r_j)\},$$
where $\text{dist}(u,r_j)$ is a distance measure between location $u$ and route $r_j$. 
The distance $\text{dist}(u,r_j)$ is calculated based on the tightness of the time window of $u$: 
\begin{align*}
\text{dist}(u,r_j) = \begin{cases}  d_{u,\text{suc}(r_j,u)} & \text{if time window of $u$ is tight, } \\
d_{u,\text{cent}(r_j)} &\text{else.}\\
\end{cases}
\end{align*}

In the instances that we considered, some customers have very tight time windows ($\sim 2\%$ of the length of the total planning horizon), and others have a time window equal to the full planning horizon.
If customer $u$ has a tight time window, we can make a well-educated guess before which customer it would be served in $r_j$, would it be added to $r_j$. We denote this successor node of $u$ in $r_j$ as $\text{suc}(r_j,u)$. It is computed by taking the first customer in $r_j$ that has an arrival time that is later than the midpoint of the time window of $u$. 
We set $\text{dist}(u,r_j)$ equal to the distance from $u$ to $\text{suc}(r_j,u)$.
Otherwise, if $u$ does not have a tight time window, we simply take the Euclidean distance from $u$ to the centroid of $r_j$, denoted by $\text{cent}(r_j)$.
The centroid of a route is defined as the arithmetic mean of its customer locations. 

Next to defining the \textsc{CreateNeighborhood} method, we need to define features that describe the potential of improvement of a neighborhood, as explained in Section \ref{ss:destroy}. 
We define several properties that describe both the routes in the neighborhood and the customers on these routes. To aggregate these features, we take the average, maximum, minimum, sum and standard deviation over the routes, for the route properties, and over the customers, for the customer properties. 
We elaborate on the type of features briefly, but give an extensive list in \ref{appendix:1}.
The first feature is the number of customers in the neighborhood. Secondly, we define properties that describe the customers in the neighborhood, like waiting time, distance contribution and closeness to other routes in the neighborhood. Thirdly, we define route properties like route distance, route duration and free capacity. Lastly, we add the distance measure $\tilde{d}(r_i, r_j)$ between all the routes in the neighborhood, as defined in Section \ref{ss:vrptw:destroy}.

\subsubsection{\textsc{Repair} Method} 
\label{ss:vrptw:repair}

The repair method considers the routes that are deleted by the destroy method and tries to create a new solution for them. 
Many repair and improvement methods are known, such as 2-opt, 3-opt, LKH-3 or exact solvers. 

In our application, a neighborhood is defined by a set of routes, so the sub-problem of finding an improvement for these routes is a VRPTW problem in itself. In fact, the number of customers and routes in this sub-problem is much smaller than in the original VRPTW. Namely, the number of vehicles in this sub-problem is equal to the number of routes that were destroyed, and the set of customers in the sub-problem is the set of customers that was destroyed. 
The repair method will find a solution for this sub-problem. 
To speed up the repair method, we can use the current solution for the destroyed routes as a warm start. 

We have seen that in practice often black-box solvers are used to repair a routing solution after it has been destroyed.
Therefore, we have also chosen to use an external VRPTW solver to solve the sub-problem defined by the deleted set of routes, namely VROOM \citep{vroom_v1.13}. VROOM is an open-source vehicle routing problem solver that is tailored towards getting high-quality solutions quickly.
By choosing such a black box solver as a repair method, we can exploit the power that this solver has on smaller instances and leverage it to solve larger instances.%
\section{Results}
\label{sec:Results}

We tested our algorithm on the $10$ R1 \citet{homberger2005two} instances with 1000 customers. To build our ML models, we needed to collect data on similar instances. Therefore we generated 100 training instances similar to the test instances. Below, we elaborate on the creation of these training instances (Section \ref{ss:results:instancegeneration}) and the subsequent data collection (Section \ref{ss:results:datacollection}). Based on the collected data, we trained a random forest classification model which we validated on a different set of data (Section \ref{ss:results:classval}). 

We tested our model with the LNS algorithm. 
Our initial findings revealed that we needed to change the data collection strategy in order for the ML model to be based on the correct data (Section \ref{ss:results:results}). 
\subsection{Instance Generation}\label{ss:results:instancegeneration}

\sloppy
Following a recommendation by 
\citet{accorsi2022Guidelines}, we used the R1 \citet{homberger2005two} instances for our experimental results. We call these instances the \emph{test instances}. There are 10 test instances, which contain 1000 customers each. The test instances only differ in the time windows. 

We want to train our ML models based on different instances from the ones we test on. Therefore, we decided to generate new training instances. Following the distribution of the original 10 test instances by \citet{homberger2005two}, we created 100 new training instances. 
These instances were created in 10 batches of 10 instances each, where each original test instance forms the basis for 10 new training instances. We call the 10 training instances that belong to one test instance its \emph{similar training instances}.
Just as in the original set of 10 test instances, the customer locations in all of the created instances are exactly the same. The only property that differs between the instances are the time windows.

The generation of the instances in a batch works as follows. 
Once for each batch, we compute for each customer what the middle of its time window will become. We do this by sampling uniformly at random over the instance horizon (to ensure feasibility of the instance, we take into consideration the time necessary to travel back from the customer to the depot when sampling the middle of the time windows) .
The 10 instances in each batch differ by the number of customers that get a restrictive time window, and the length that this restrictive time window will be. 
All customers with a non-restrictive time window can be served during the whole horizon. 
The first 4 instances in a batch have a time window length of 10, and the fraction of customers with a restrictive time window is 100\%, 75\%, 50\% and 25\%, respectively.
The same holds for the next 4 instances in the batch, but with a time window length of 30.
In the last two instances of a batch all customers have a restrictive time window, which length is sampled from a normal distribution with means 60 and 120 and standard deviation 20 and 30 respectively. 
The details of this instance generation are summarized in Table \ref{tbl:instance_generation}.

\begin{table}[ht]
\small
\begin{tabular}{lrrrrrrrrrr}
\toprule
%            & 1  & 2    & 3   & 4    & 5  & 6    & 7   & 8    & 9           & 10           \\ \hline
Test Instance Name & R1\_10\_1 & R1\_10\_2 & R1\_10\_3 & R1\_10\_4 & R1\_10\_5 \\ \midrule %& R1\_10\_6 & R1\_10\_7 & R1\_10\_8 & R1\_10\_9 & R1\_10\_10 \\ \hline
TW Length   & 10 & 10   & 10  & 10   & 30 \\ %%& 30   & 30  & 30   & $\mathcal{N}(60,20)$ & $\mathcal{N}(120,30)$ \\
TW Fraction & 100\%  & 75\% & 50\% & 25\% & 100\% \\ \toprule\toprule % & 75\% & 50\% & 25\% & 100\%           & 100\%           
Test Instance Name & R1\_10\_6 & R1\_10\_7 & R1\_10\_8 & R1\_10\_9 & R1\_10\_10 \\ \midrule
TW Length   & 30   & 30  & 30   & $\mathcal{N}(60,20)$ & $\mathcal{N}(120,30)$ \\
TW Fraction & 75\% & 50\% & 25\% & 100\%           & 100\%          \\ \bottomrule
\end{tabular}
\caption{Time window length (or distribution) and fraction of customers with time window for 10 test instances used during instance generation.}
\label{tbl:instance_generation}
\end{table}

\subsection{Data Collection}\label{ss:results:datacollection}

In order to train an ML model, we collected data on these generated instances, as explained in Section \ref{ss:destroy}.
For each of the 100 training instances, we did 10 runs of data collection of 500 iterations each. 
We used the random data collection strategy to build an ML model, ML1, and followed our Guidelines for building the consecutive ML models, ML2, ML3, ML4 and ML5, based on more data. 
The sample that we saved for each neighborhood in an iteration, defined by $(\samplex_{ij}, \sampley_{ij})$ (Section \ref{ss:destroy}) consists of $\samplex_{ij}$, the features describing the neighborhood (Section \ref{ss:vrptw:destroy}), and $\sampley_{ij}$, the improvement that VROOM is able to find for this neighborhood. 
This improvement is defined as the difference in total travel distance between the routes in the neighborhood before VROOM optimizes them and after optimization. 

There are 10 \emph{similar training instances} for each test instance. We did 10 training runs of 500 iterations, during which 10 samples were stored in each iteration. This gave $10\cdot 10\cdot 500\cdot 10=500,000$ samples for each test instance, on which we based the random forest model for ML1. Each consecutive created ML model (ML2, ML3, ML4 and ML5) was based on 500,000 more samples.

\subsection{Benchmarking Algorithms}

To test our ML models, we defined two algorithms to compare with: the oracle model and the random model. Both of these algorithms follow the \textsc{Large Neighborhood Search} algorithm, {and like \MLlns, they create \nhperit\ neighborhoods in each iteration. However, the benchmarking algorithms differ in how they select one of these neighborhoods.}

The random model selects one of the created 10 neighborhoods at random. In fact, the random model mimics the situation in which only one neighborhood is created per iteration, since the other nine neighborhoods may be regarded as never created. We hope our ML algorithm will be able to beat this random model. 

The second benchmark is the oracle model, which selects the neighborhood that would yield the highest improvement if it was destroyed and repaired. The oracle model can be seen as an ML model with perfect predictions; or, alternatively, it represents the potential that the best (though fictitious) \MLlns\ algorithm could give. To acquire this full knowledge the oracle needs to compute the destroy and repair step for each of the 10 neighborhoods. As a consequence, it is a very slow and expensive procedure. In fact, for most practical applications it is even infeasible to compute it.
But it provides an insightful and strong benchmarking algorithm for our experiments.

\subsection{Classification and Validation}\label{ss:results:classval}

The 500,000 collected samples were partitioned into a training (60\%) and validation (40\%) set. 
After the collection, we labeled the samples with an improvement threshold of 0, meaning that a sample belongs to the positive class if it has a positive improvement, and to the negative class otherwise, as explained in Section \ref{ss:destroy}. This resulted in a heavily unbalanced dataset with, e.g., only 11\% of training samples being positive (for the first test instance). Therefore, after applying a standard scalar to normalize the data, we balanced the sample weights and trained a random forest model. For each of the test instances, we created a random forest like this. We call these models \emph{ML1}.

We validated our ML1 model on two different sets of samples. The first set is the \emph{validation samples}, set apart earlier, based on the similar training instances. Next to that, we created a set of samples, based on the test instance, which we call the \emph{test samples}. We created the test samples by running the data collection with the random data collection strategy on the test instance and obtained 50,000 samples coming from 5,000 iterations. 

The validation of an Algorithm ALG works as follows. Recall that a sample is denoted by $(\samplex_{ij}, \sampley_{ij})$, where $i$ denotes the iteration and $j \in \{1, \ldots, \nhperit \}$ is the index for the neighborhood in iteration $i$. 
In each iteration $i$, we let ALG decide which of the \nhperit\ neighborhoods is the best, and denote its index with $j^{ALG}$. Then for each iteration $i$ we save $y_{i,j^{ALG}}$ and compute the average over all iterations, this value indicates the average improvement of the neighborhoods that are considered best by ALG. Moreover, we check in how many of the iterations $y_{i,j^{ALG}}$ was a strictly positive improvement. This value indicates the fraction of iterations in which ALG would be able to find an improvement.

The results for the validation samples are given in Table \ref{tab:instanceLikeR1_train} and the results for the test samples are given in Table \ref{tab:instanceLikeR1_test}.%
\footnote{For this validation, we only considered the iterations in which there was at least one improving neighborhood.}
%\gs{footnote after punctuation}
Table \ref{tab:instanceLikeR1_train} and Table \ref{tab:instanceLikeR1_test} show the oracle model, the random model and the ML1 model. 

For the validation samples, we see that the ML1 model is able to identify significantly more improving neighborhoods than the random model (44\% against 18\%). Also, the average improvement is about three times higher for the ML model. For the test samples, we see that the ML model is able to find an improvement in 28\% of the iterations, while the random model only finds an improvement in 20\% of the cases. Moreover, we see that the average improvement of the ML model (4.7) is about 46\% higher than that of the random model (3.2). Even though we lose some prediction power if we start predicting for samples from the test set, we are still able to outperform random. 
\begin{table}[ht]
    \centering
    \begin{tabular}{lrrr} \toprule
         Neighborhood Selection Model& Oracle & Random & ML1 \\ \midrule %\cmidrule(r){2-2}\cmidrule(r){3-3}\cmidrule(r){4-4}
         Average true rank of best prediction & 1 & 2.53 & 2.1\\
         Average fraction of improving neighborhoods & 100\% & 18\% & 44\% \\
         Average improvement & 21.7 & 3.0 & 8.8 \\ \bottomrule
    \end{tabular}
    \caption{Validation of ML1 on validation samples for three neighborhood selection models}
    \label{tab:instanceLikeR1_train}
\end{table}

\begin{table}[ht]
    \centering
    \begin{tabular}{lrrr} \toprule
         Neighborhood Selection Model& Oracle & Random & ML1 \\ \midrule %\cmidrule(r){2-2}\cmidrule(r){3-3}\cmidrule(r){4-4}
         Average true rank of best prediction & 1 & 2.67 & 2.52\\
         Average fraction of improving neighborhoods & 100\% & 20\% & 28\% \\
         Average improvement & 21.3 & 3.2 & 4.7 \\ \bottomrule
    \end{tabular}
    \caption{Validation of ML1 on test samples for three neighborhood selection models}
    \label{tab:instanceLikeR1_test}
\end{table}

\subsection{Algorithm Results}\label{ss:results:results}

After the validations, we checked how our ML1 performed in the LNS algorithm introduced in Section \ref{Destroy-learn}. We tested this for the 10 test instances and started the optimization with an initial feasible solution created by VROOM. 
All of the tests ran for 500 iterations. 

Figure \ref{fig:result-figures} shows the decrease of the total distance during the optimization runs of LNS with the ML1 model compared to LNS with the random model and LNS with the oracle model. Table \ref{tab:results} and \ref{tab:results2} show the total distance after 500 and 200 iterations, respectively. All shown results are averages over 10 runs. Next to these average distances, the tables show the best-known solution for these instances, as recorded by \citep{Sintef}. 

We compare our ML algorithms with the oracle and the random model by using a gap measure, which indicates how far off the results with ML are from the best-case algorithm, which is the oracle. A gap of 100\% means the model is as bad as the random model, a gap of 0\% means the model is as good as the oracle model. 

Figure \ref{fig:result-figures}, Table \ref{tab:results} and \ref{tab:results2} show that the quality of the solution increases during optimization since the total distance of the solution decreases. 
The oracle (which takes the most time and effort per iteration) has the lowest total distance after 500 iterations, as expected. Unfortunately, we see for most of the instances that the ML1 model is not able to improve over the random model. This means that for those instances, following the recommendation by the ML1 model is worse than following random choices. The average gap of 110.04\% also shows that ML1 is not able to beat random. 

At first sight, this seems a contradiction to the validation that was shown in the previous section. However, a more in-depth analysis of our experiments reveals that,
because we follow the choices based on ML, 
we enter a part of the solution space that was not encountered during the data collection. Since the ML model was not trained on this part of the solution space, it does not perform well here.

\begin{figure}[ht]
\centering
\begin{subfigure}[b]{0.45\textwidth}
    \centering
    \includegraphics[width=0.9\hsize]{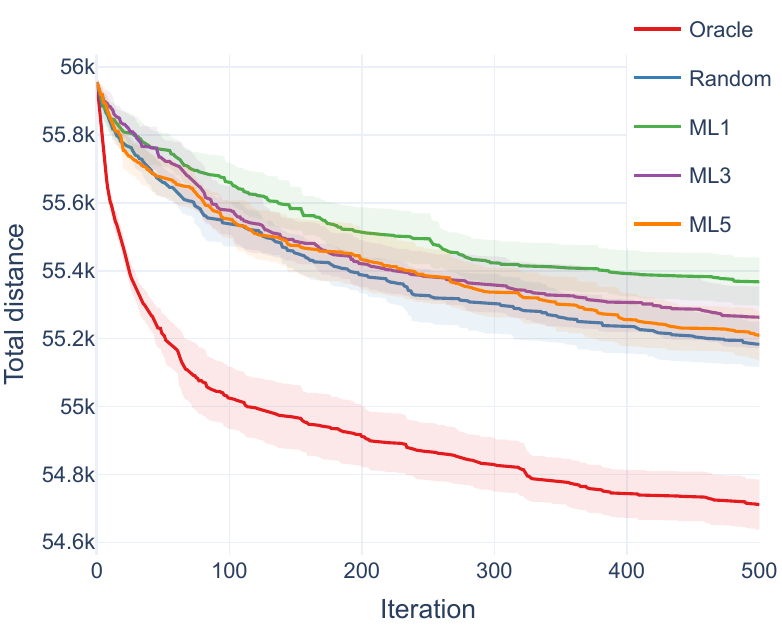}
    \caption{Total Distance for instance R1\_10\_1}
    \label{fig:resultfig1}
\end{subfigure}
    \hfil
\begin{subfigure}[b]{0.45\textwidth}
    \centering
    \includegraphics[width=0.9\hsize]{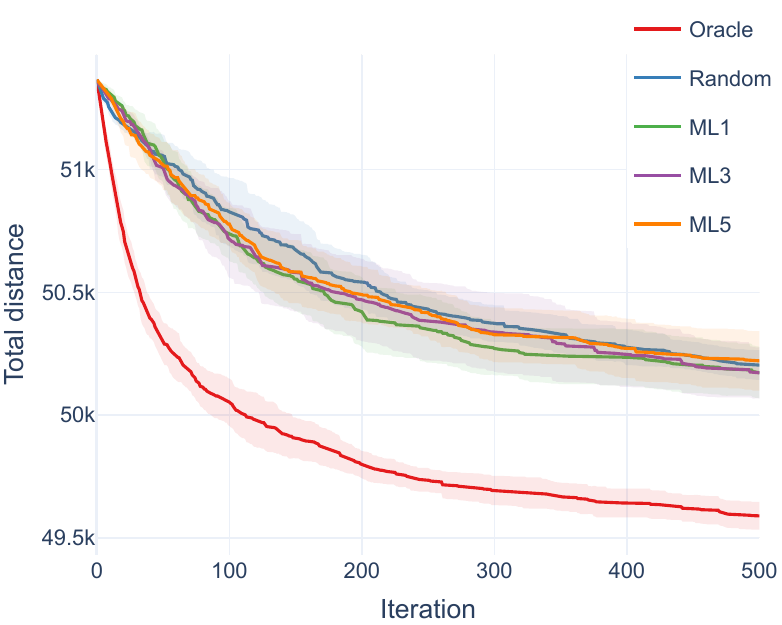}
    \caption{Total Distance for instance R1\_10\_2}
    \label{fig:resultfig2}
\end{subfigure}
%\vspace{1cm}

\begin{subfigure}[b]{0.45\textwidth}
    \centering
    \includegraphics[width=0.9\hsize]{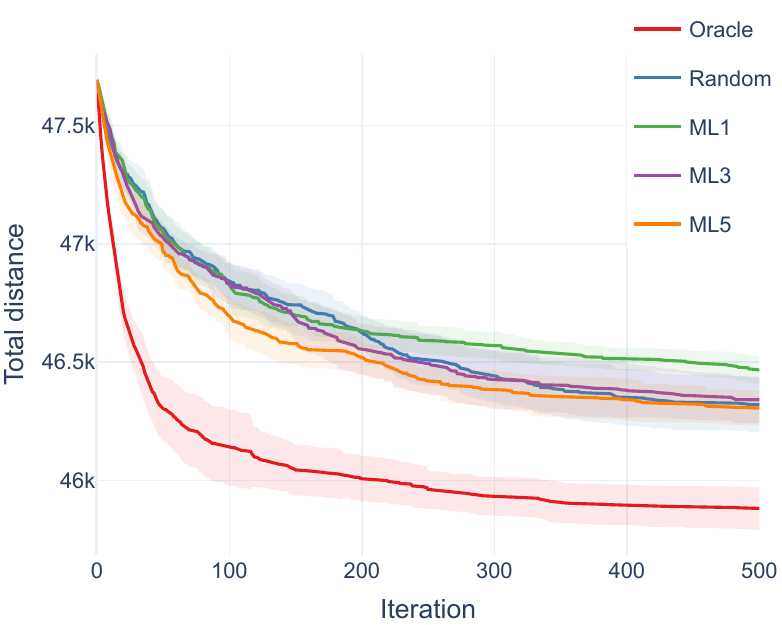}
    \caption{Total Distance for instance R1\_10\_3}
    \label{fig:resultfig3}
\end{subfigure}
    \hfil
\begin{subfigure}[b]{0.45\textwidth}
    \centering
    \includegraphics[width=0.9\hsize]{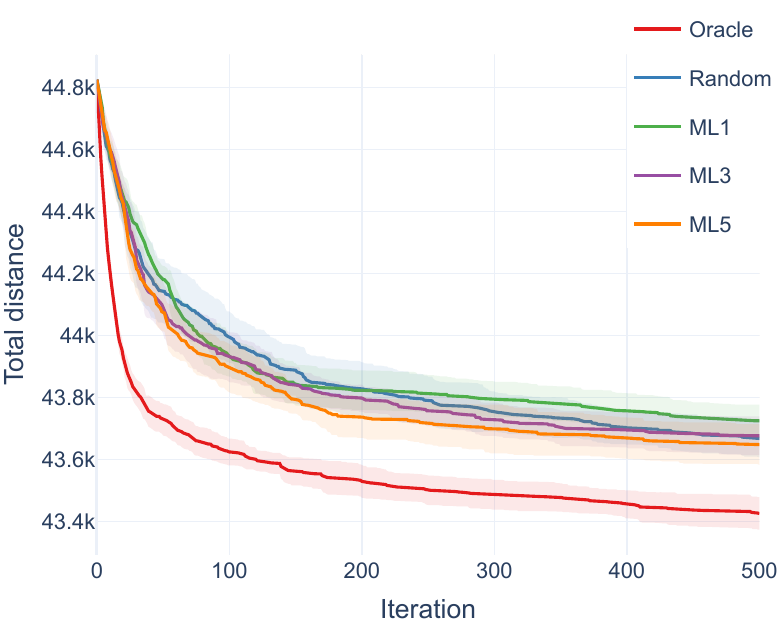}
    \caption{Total Distance for instance R1\_10\_4}
    \label{fig:resultfig4}
\end{subfigure}
%\vspace{1cm}

\begin{subfigure}[b]{0.45\textwidth}
    \centering
    \includegraphics[width=0.9\hsize]{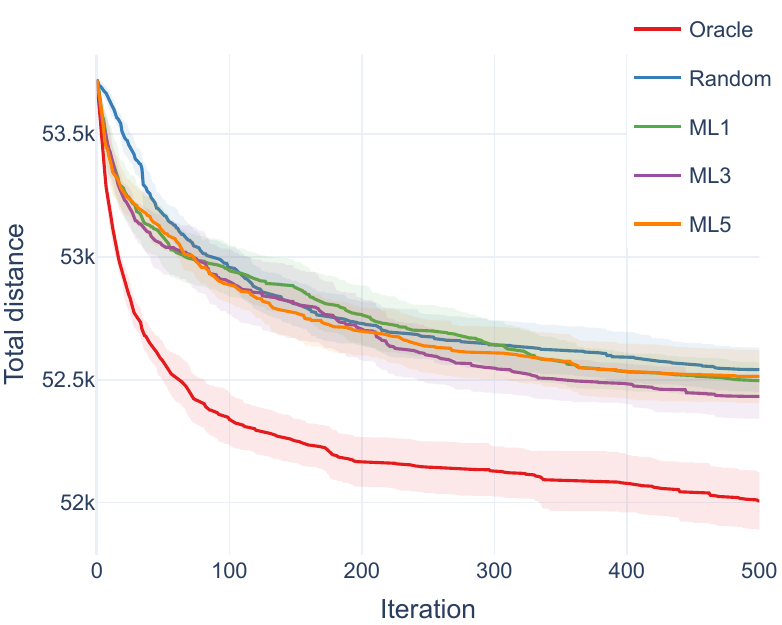}
    \caption{Total Distance for instance R1\_10\_5}
    \label{fig:resultfig5}
\end{subfigure}
    \hfil
\begin{subfigure}[b]{0.45\textwidth}
    \centering
    \includegraphics[width=0.9\hsize]{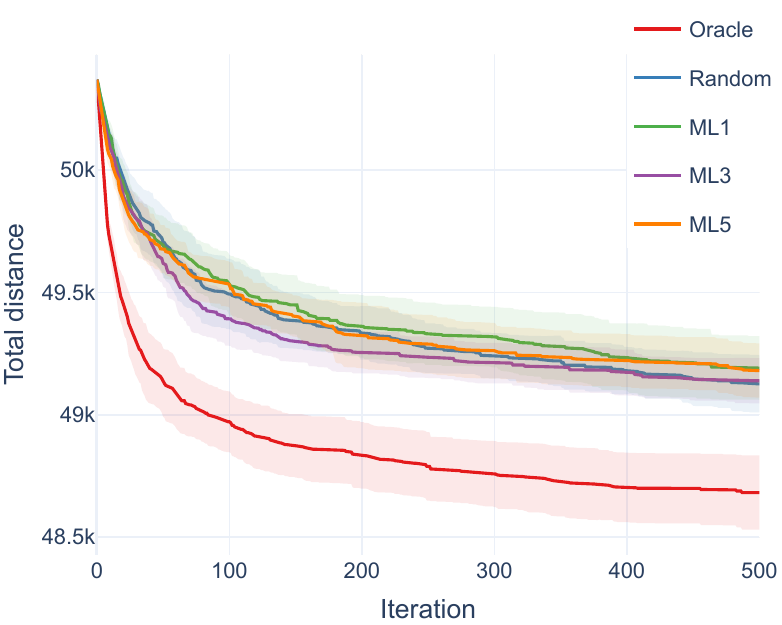}
    \caption{Total Distance for instance R1\_10\_6}
    \label{fig:resultfig6}
\end{subfigure}
\caption{Total distance for 10 test instances for the oracle model, the random model, ML1 model, ML3 model and ML5 model}
\end{figure}

\begin{figure}\ContinuedFloat
\centering
\begin{subfigure}[b]{0.45\textwidth}
    \centering
    \includegraphics[width=0.9\hsize]{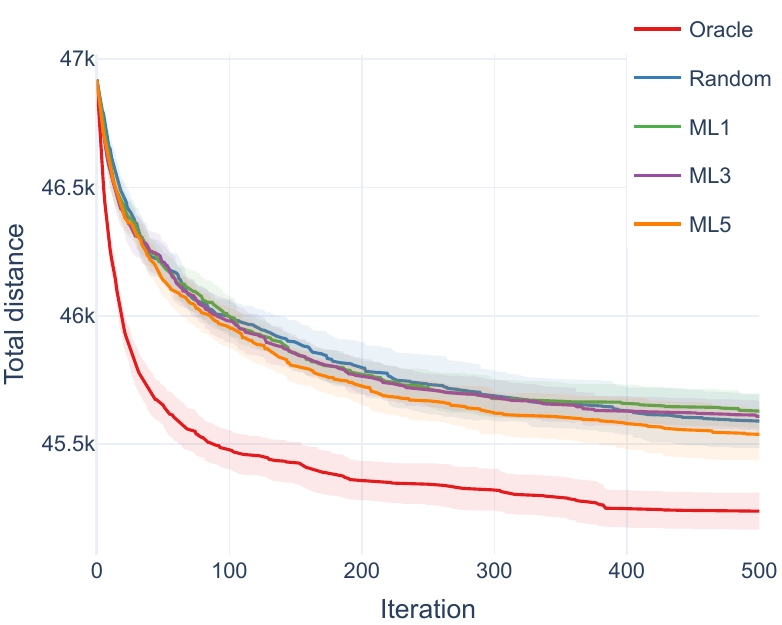}
    \caption{Total Distance for instance R1\_10\_7}
    \label{fig:resultfig7}
\end{subfigure}
    \hfil
\begin{subfigure}[b]{0.45\textwidth}
    \centering
    \includegraphics[width=0.9\hsize]{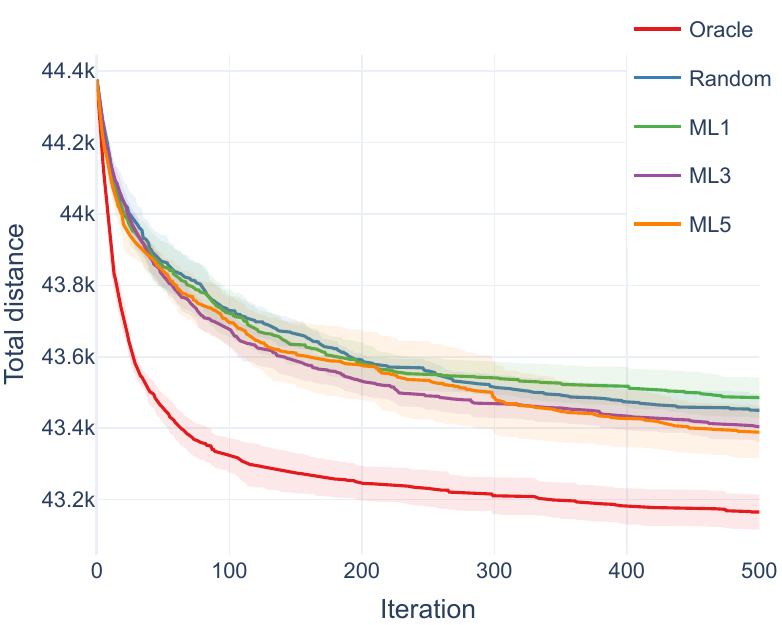}
    \caption{Total Distance for instance R1\_10\_8}
    \label{fig:resultfig8}
\end{subfigure}
%\vspace{1cm}

\begin{subfigure}[b]{0.45\textwidth}
    \centering
    \includegraphics[width=0.9\hsize]{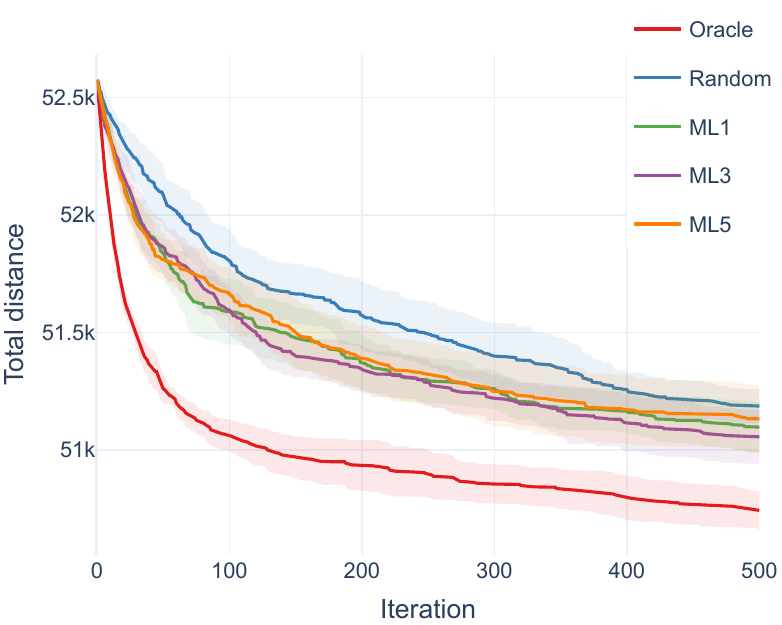}
    \caption{Total Distance for instance R1\_10\_9}
    \label{fig:resultfig9}
\end{subfigure}
    \hfil
\begin{subfigure}[b]{0.45\textwidth}
    \centering
    \includegraphics[width=0.9\hsize]{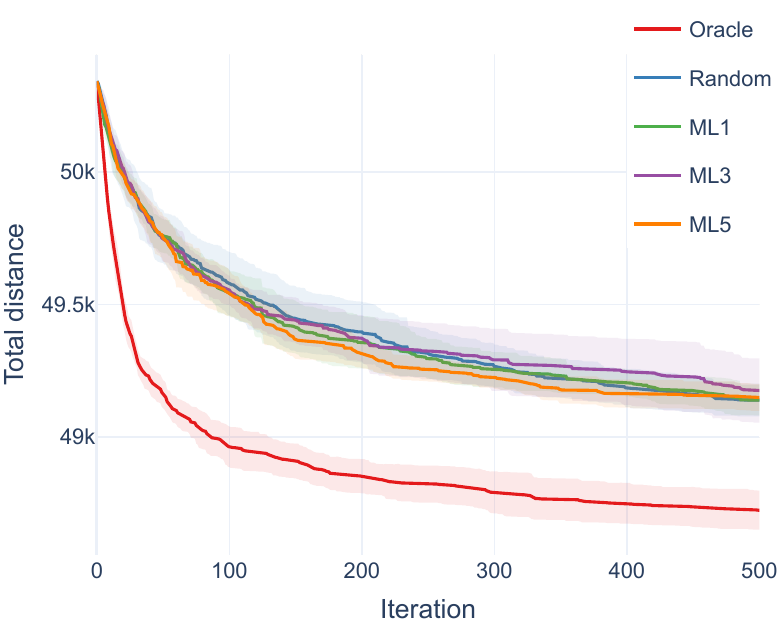}
    \caption{Total Distance for instance R1\_10\_10}
    \label{fig:resultfig10}
\end{subfigure}
    \caption{Total distance for 10 test instances for the oracle model, the random model, ML1 model, ML3 model and ML5 model (cont.)}
    \label{fig:result-figures}
\end{figure}

Therefore we believe that it is crucial to do multiple runs of data collection, as stated in our Guidelines (Section \ref{ss:destroy}). In the second (third/fourth/fifth) run of data collection, instead of using the \emph{random} data collection strategy, we used the ML1 (ML2/ML3/ML4) model during data collection. That is, the ML1 (ML2/ML3/ML4) model decides which of the neighborhoods suggested by VROOM should be implemented. 
This resulted in an expanded dataset of samples. On this expanded dataset, consisting of all data collected in previous training rounds, we trained a new random forest model, denoted as \emph{ML2} (ML3/ML4/ML5).
{Table \ref{tab:results} and Table \ref{tab:results2} show the results for ML5, Table \ref{tab:resultsfull} and Table \ref{tab:results2full} in \ref{app:fullresults} show the extended results, including ML3.}
The gap improved to 97.10\% for ML3, and 95.74\% for ML5.
This significant improvement compared to ML1 shows the importance of performing multiple rounds of data collection.
After 200 iterations, we see a larger improvement compared to the random algorithm, with an average gap of 88.20\%. 
This shows that mainly in the first half of the optimization run, we have learned to do better than random.

\begin{table}[ht]
\scriptsize
    \centering
\begin{tabular}{@{}lrrrrrrr@{}} \toprule
&&\multicolumn{1}{c}{Oracle}
&\multicolumn{1}{c}{Random}
&\multicolumn{2}{c}{ML1} 
&\multicolumn{2}{c}{ML5}\\ \cmidrule(l){3-3}\cmidrule(l){4-4}\cmidrule(l){5-6}\cmidrule(l){7-8}
Instance    & BKS     & Avg     & Avg     & Avg     & Gap      & Avg     & Gap\\ \midrule
R1\_10\_1   & 53380.2 & 54711.2 & 55183.8 & 55367.5 & 138.86\% & 55206.9 & 104.88\%\\
R1\_10\_2   & 48261.0 & 49589.3 & 50203.6 & 50172.5 & 94.94\%  & 50221.0 & 102.84\%\\
R1\_10\_3   & 44720.9 & 45881.8 & 46320.1 & 46467.1 & 133.55\% & 46305.6 & 96.71\%\\
R1\_10\_4   & 42463.7 & 43426.6 & 43668.2 & 43725.2 & 123.59\% & 43648.6 & 91.87\%\\
R1\_10\_5   & 50452.9 & 52007.1 & 52541.1 & 52496.4 & 91.62\%  & 52513.9 & 94.90\%\\
R1\_10\_6   & 46966.5 & 48682.6 & 49126.6 & 49191.3 & 114.57\% & 49181.0 & 112.26\%\\
R1\_10\_7   & 43978.7 & 45238.5 & 45589.6 & 45628.4 & 111.07\% & 45537.9 & 85.29\%\\
R1\_10\_8   & 42291.7 & 43165.2 & 43450.1 & 43485.5 & 112.44\% & 43388.8 & 78.47\%\\
R1\_10\_9   & 49208.1 & 50743.9 & 51187.7 & 51098.2 & 79.82\%  & 51133.3 & 87.75\%\\
R1\_10\_10  & 47407.2 & 48724.0 & 49139.5 & 49139.4 & 99.98\%  & 49149.4 & 102.40\%\\
\midrule
Average & 46913.1 & 48217.0 & 48641.0 & 48677.1 & 110.04\%  & 48628.6 & 95.74\%\\ \bottomrule
\end{tabular}
\caption{Total distance after 500 iterations of the oracle model, random model, ML1 model and ML5 model. Also the total distance of the best-known solution (BKS). Recall that the gaps of the oracle and random models are defined as $0\%$ and $100\%$, respectively.}
    \label{tab:results}
\end{table}

\begin{table}[ht]
\scriptsize
    \centering
\begin{tabular}{@{}lrrrrrrr@{}} \toprule
&&\multicolumn{1}{c}{Oracle}
&\multicolumn{1}{c}{Random}
&\multicolumn{2}{c}{ML1} 
&\multicolumn{2}{c}{ML5}\\ \cmidrule(l){3-3}\cmidrule(l){4-4}\cmidrule(l){5-6}\cmidrule(l){7-8}
Instance    & BKS     & Avg     & Avg     & Avg     & Gap      & Avg     & Gap\\ \midrule
R1\_10\_1   & 53380.2 & 54911.8 & 55388.8 & 55512.3 & 125.89\% & 55431.8 & 109.01\%\\
R1\_10\_2   & 48261.0 & 49798.1 & 50541.5 & 50420.8 & 83.77\%  & 50491.1 & 93.22\%\\
R1\_10\_3   & 44720.9 & 46007.4 & 46623.0 & 46634.7 & 101.90\% & 46518.7 & 83.06\%\\
R1\_10\_4   & 42463.7 & 43529.9 & 43827.8 & 43824.0 & 98.75\%  & 43737.1 & 69.55\%\\
R1\_10\_5   & 50452.9 & 52166.3 & 52729.2 & 52765.2 & 106.40\% & 52696.7 & 94.23\%\\
R1\_10\_6   & 46966.5 & 48834.9 & 49335.0 & 49361.6 & 105.32\% & 49324.4 & 97.88\%\\
R1\_10\_7   & 43978.7 & 45357.2 & 45797.3 & 45770.6 & 93.94\%  & 45724.7 & 83.50\%\\
R1\_10\_8   & 42291.7 & 43245.9 & 43591.1 & 43582.6 & 97.53\%  & 43576.2 & 95.66\%\\
R1\_10\_9   & 49208.1 & 50935.6 & 51571.6 & 51372.7 & 68.72\%  & 51390.6 & 71.54\%\\
R1\_10\_10  & 47407.2 & 48852.1 & 49396.4 & 49357.5 & 92.86\%  & 49311.2 & 84.35\%\\
\midrule                  
Average     & 46913.1 & 48363.9 & 48880.2 & 48860.2 & 97.51\%  & 48820.2 & 88.20\%\\
\bottomrule
\end{tabular}
    \caption{Total distance after 200 iterations of the oracle model, random model, ML1 model and ML5 model. Also the total distance of the best-known solution (BKS). 
    Recall that the gaps of the oracle and random models are defined as $0\%$ and $100\%$, respectively.}
    \label{tab:results2}
\end{table}

\section{Conclusion and Discussion}

This research was inspired by experimental results that we obtained for a real-world application, where we used our LENS approach to enhance the neighborhood selection
of an LNS algorithm. 

The results described in this paper were obtained through a similar approach using publicly available synthetic VRPTW datasets.
On these VRPTW instances, our \MLlns\ technique gives a solution with a cost that is 11.8\% smaller after 200 iterations and 4.26\% smaller after 500 iterations, compared to a solution obtained with random neighborhood selection. The results that we obtained on the real-world application were even more significant. 

We believe that this is mainly caused by the underlying baseline algorithm to which the new destroy method is added. 
In the real-world application, the baseline algorithm is a sophisticated and very well-working algorithm, and there are many neighborhoods in an iteration that yield an improvement. In our baseline algorithm, however, this number is much lower. It seems that the destroy method with learning benefits from a large number of improving neighborhoods since it is easier to identify large improvements if there are many promising neighborhoods available in the candidate set.
In our baseline algorithm, however, the number of promising neighborhoods per iteration is much smaller and therefore it is also harder to select a good one. Nevertheless, we still obtain the improvements of 11.8\% after 200 iterations and 4.26\% after 500 iterations. 

Given that LNS is a universal approach that is broadly applicable and easy to implement, we believe that our LENS approach is a promising way to improve the quality of the computed solutions through the help of ML techniques. The actual magnitude of improvement might be application-specific though and, in particular, depends on the quality of the created neighborhoods.

\bigskip\bigskip\noindent
\textbf{Acknowledgements.} This research is partly funded by DELMIA, a Dassault Systèmes brand. DELMIA helps industries and services to collaborate, model, optimize, and perform their operations.

\bibliographystyle{elsarticle-num-names-sorted} 
\bibliography{ms.bib}

\appendix

\newpage
\section{ML model features}
\label{appendix:1}

We define several properties that describe both the routes in the neighborhood and the customers on these routes. To aggregate these features, we take the average, maximum, minimum, sum and standard deviation over the routes, for the route properties, and over the customers, for the customer properties. 
The first feature is the number of customers in the neighborhood.
The following properties describe the customers in the neighborhood: 
\begin{itemize}
    %\item lateness. The lateness of a customer is zero if the arrival time is before the end of the time window, otherwise it is the positive difference between the arrival time and the end of the time window.
    \item Waiting time: The waiting time of a customer is zero if the arrival time is after the start of the time window, otherwise it is the positive difference between the arrival time and the start of the time window. 
    \item Closeness: The closeness of customer $c$ is the minimum of the distances between $c$ and all customers in other routes (than the one that contains $c$) in the neighborhood.
    \item Temporal closeness: Like closeness, but defined with temporal distance instead of the Euclidean distance. Temporal distance is defined as the sum of the Euclidean distance and the time window difference. The time window difference indicates how compatible the time windows of two shipments are. The time window difference between two clients is the minimum waiting time caused by serving the clients directly after each other, in any order, if feasible. If the time windows make it infeasible to serve the clients after each other, the time window difference is set to a large penalty value.
%\footnote{
%Let $\textsc{wait}(c_1, c_2)$ be the waiting time at $c_2$ if it is served directly after $c_1$ and let $\textsc{lateness}(c_1, c_2)$ be the time that the vehicle is late at $c_2$ if it is served directly after $c_1$. Clearly, at most one of $\textsc{wait}(c_1, c_2)$ and $\textsc{lateness}(c_1,c_2)$ can be non-zero. The time window difference between $c_1$ and $c_2$ is computed as follows: $\textsc{TimeWindowDifference}(c_1,c_2) = \min( \textsc{wait}(c_1, c_2) + \textsc{lateness}(c_1,c_2), \textsc{wait}(c_2,c_1) + \textsc{lateness}(c_2,c_1)$.}
    \item Centroid closeness: Compute the distance between the customer and the centroid of each other route in the neighborhood. The centroid closeness is the minimum over these distances.
    is the minimum distance from $c$ to the centroid of another route in the neighborhood.
    \item Distance contribution: The difference between the length of the route in which the customer lies and the length of the route if the customer would be removed. 
    \item Time window length: The length of the time window of the customer.
    \item Distance to the depot: The distance from the customer to the depot.
    \item Load: The demand of the customer.
    \item Minimum greedy addition cost: For each other route in the neighborhood, the greedy addition cost is how much the distance increases if the customer is added to this route. The minimum greedy addition cost is the minimum of these increases over all the other routes in the neighborhood. The greedy addition is calculated greedily, since only one possible location in the other route is tested: before the first customer that has a time window that starts later.
    %The minimum over the increase in distance if the customer is added to another route in the neighborhood. The addition is greedy since we test only one position in the other route, based on the start of the time windows. 
    \item Maximum gain: For each other route in the neighborhood, compute the gain of adding the customer to the other route by subtracting the greedy addition cost in the other route from the distance contribution in the customer's route.
    The maximum gain is the maximum of these gains over all other routes. 
    \item Possible delay: The difference between the end of the time window and the current arrival time.
\end{itemize}
Moreover, we compute the following features which describe the routes in the neighborhood:
\begin{itemize}
    \item Route distance: The distance of the route.
    \item Average route distance: The distance of the route, divided by the number of customers in the route. 
    \item Empty distance: The distance that the vehicle drives after the last delivery.
    \item Worst case distance fraction: The distance of the route divided by the distance of a so-called worst-case solution. The worst-case solution is computed by traveling back to the depot after each customer in the route.
    \item Route duration: The travel time plus the waiting time plus the service time. 
    \item Average route duration: The route duration divided by the number of customers in the route. 
    \item Idle time: The total time the vehicle is not traveling nor servicing.
    \item Free capacity: The free capacity in the vehicle.
    \item Fitting candidates: The number of customers in the other routes in the neighborhood that have a demand that is smaller than the free capacity of the vehicle. 
    \item Expected number fitting candidates: The free capacity of the vehicle divided by the average demand of the customers on the other routes in the neighborhood.
    %Computed by dividing the free capacity of the vehicle by the average demand of the customers who are not on this vehicle.
\end{itemize}
Lastly, we add the handcrafted distance measure 
$\tilde{d}(r_i, r_j)$ between all the routes in the neighborhood, defined in Section \ref{ss:vrptw:destroy}.

\newpage

\section{Extended Results}\label{app:fullresults}%
\begin{table}[!htbp]
\centering
\rotatebox{90}{%
\begin{minipage}{0.9\textheight}
\captionsetup{width=5.5in}
\scriptsize
\begin{tabular}{@{}lrrrrrrrrrrr@{}} \toprule
&&\multicolumn{2}{c}{Oracle}
&\multicolumn{2}{c}{Random}
&\multicolumn{2}{c}{ML1} 
&\multicolumn{2}{c}{ML3} 
&\multicolumn{2}{c}{ML5}\\ \cmidrule(l){3-4} \cmidrule(l){5-6}\cmidrule(l){7-8}\cmidrule(l){9-10}\cmidrule(l){11-12}
Instance & BKS & Avg & Gap 
& Avg & Gap %& Avg & Gap  & Avg & Gap 
& Avg & Gap & Avg & Gap & Avg & Gap\\ \midrule
R1\_10\_1 & 53380.2 & 54711.2 & 0.00\% & 55183.8 & 100.00\% & 55367.5 & 138.86\% & 55263.2 & 116.80\% & 55206.9 & 104.88\%\\
R1\_10\_2 & 48261.0 & 49589.3 & 0.00\% & 50203.6 & 100.00\% & 50172.5 & 94.94\% & 50172.0 & 94.85\% & 50221.0 & 102.84\%\\
R1\_10\_3 & 44720.9 & 45881.8 & 0.00\% & 46320.1 & 100.00\% & 46467.1 & 133.55\% & 46341.3 & 104.85\% & 46305.6 & 96.71\%\\
R1\_10\_4 & 42463.7 & 43426.6 & 0.00\% & 43668.2 & 100.00\% & 43725.2 & 123.59\% & 43676.8 & 103.55\% & 43648.6 & 91.87\%\\
R1\_10\_5 & 50452.9 & 52007.1 & 0.00\% & 52541.1 & 100.00\% & 52496.4 & 91.62\% & 52432.3 & 79.62\% & 52513.9 & 94.90\%\\
R1\_10\_6 & 46966.5 & 48682.6 & 0.00\% & 49126.6 & 100.00\% & 49191.3 & 114.57\% & 49139.6 & 102.93\% & 49181.0 & 112.26\%\\
R1\_10\_7 & 43978.7 & 45238.5 & 0.00\% & 45589.6 & 100.00\% & 45628.4 & 111.07\% & 45608.2 & 105.30\% & 45537.9 & 85.29\%\\
R1\_10\_8 & 42291.7 & 43165.2 & 0.00\% & 43450.1 & 100.00\% & 43485.5 & 112.44\% & 43403.8 & 83.77\% & 43388.8 & 78.47\%\\
R1\_10\_9 & 49208.1 & 50743.9 & 0.00\% & 51187.7 & 100.00\% & 51098.2 & 79.82\% & 51057.0 & 70.54\% & 51133.3 & 87.75\%\\
R1\_10\_10 & 47407.2 & 48724.0 & 0.00\% & 49139.5 & 100.00\% & 49139.4 & 99.98\% & 49175.8 & 108.74\% & 49149.4 & 102.40\%\\
\midrule
Average & 46913.1 & 48217.0 & 0.00\% & 48641.0 & 100.00\% & 48677.1 & 110.04\% & 48627.0 & 97.10\% & 48628.6 & 95.74\%\\ \bottomrule
\end{tabular}
\caption{Total distance after 500 iterations of the oracle model, random model, ML1 model, ML3 model and ML5 model. Also the total distance of the best-known solution (BKS).}
\label{tab:resultsfull}
\end{minipage}}
\end{table}
\begin{table}[!htbp]
\centering
\rotatebox{90}{%
\begin{minipage}{0.9\textheight}
\captionsetup{width=5.5in}
\scriptsize
\begin{tabular}{@{}lrrrrrrrrrrr@{}} \toprule
&&\multicolumn{2}{c}{Oracle}
&\multicolumn{2}{c}{Random}
&\multicolumn{2}{c}{ML1} 
&\multicolumn{2}{c}{ML3} 
&\multicolumn{2}{c}{ML5}\\ \cmidrule(l){3-4} \cmidrule(l){5-6}\cmidrule(l){7-8}\cmidrule(l){9-10}\cmidrule(l){11-12}
Instance & BKS & Avg & Gap 
& Avg & Gap %& Avg & Gap  & Avg & Gap 
& Avg & Gap & Avg & Gap & Avg & Gap\\ \midrule
R1\_10\_1   & 53380.2 & 54911.8 & 0.00\% & 55388.8 & 100.00\% & 55512.3 & 125.89\% & 55421.2 & 106.79\%& 55431.8 & 109.01\%\\
R1\_10\_2   & 48261.0 & 49798.1 & 0.00\% & 50541.5 & 100.00\% & 50420.8 & 83.77\%  & 50468.7 & 90.22\% & 50491.1 & 93.22\%\\
R1\_10\_3   & 44720.9 & 46007.4 & 0.00\% & 46623.0 & 100.00\% & 46634.7 & 101.90\% & 46553.7 & 88.75\% & 46518.7 & 83.06\%\\
R1\_10\_4   & 42463.7 & 43529.9 & 0.00\% & 43827.8 & 100.00\% & 43824.0 & 98.75\%  & 43798.6 & 90.20\% & 43737.1 & 69.55\%\\
R1\_10\_5   & 50452.9 & 52166.3 & 0.00\% & 52729.2 & 100.00\% & 52765.2 & 106.40\% & 52705.1 & 95.71\% & 52696.7 & 94.23\%\\
R1\_10\_6   & 46966.5 & 48834.9 & 0.00\% & 49335.0 & 100.00\% & 49361.6 & 105.32\% & 49255.3 & 84.06\% & 49324.4 & 97.88\%\\
R1\_10\_7   & 43978.7 & 45357.2 & 0.00\% & 45797.3 & 100.00\% & 45770.6 & 93.94\%  & 45763.9 & 92.40\% & 45724.7 & 83.50\%\\
R1\_10\_8   & 42291.7 & 43245.9 & 0.00\% & 43591.1 & 100.00\% & 43582.6 & 97.53\%  & 43531.4 & 82.70\% & 43576.2 & 95.66\%\\
R1\_10\_9   & 49208.1 & 50935.6 & 0.00\% & 51571.6 & 100.00\% & 51372.7 & 68.72\%  & 51344.0 & 64.22\% & 51390.6 & 71.54\%\\
R1\_10\_10  & 47407.2 & 48852.1 & 0.00\% & 49396.4 & 100.00\% & 49357.5 & 92.86\%  & 49372.8 & 95.66\% & 49311.2 & 84.35\%\\
\midrule
Average     & 46913.1 & 48363.9 & 0.00\% & 48880.2 & 100.00\% & 48860.2 & 97.51\%  & 48821.5 & 89.07\% & 48820.2 & 88.20\%\\
\bottomrule
\end{tabular}
    \caption{Total distance after 200 iterations of the oracle model, random model, ML1 model, ML3 model and ML5 model. Also the total distance of the best-known solution (BKS).}
    \label{tab:results2full}
\end{minipage}}
\end{table}

\end{document}